\documentclass[conference]{IEEEtran}
\IEEEoverridecommandlockouts

\usepackage{cite}
\usepackage{amsmath,amssymb,amsfonts}
\usepackage{algorithmic}
\usepackage{graphicx}
\usepackage{textcomp}
\usepackage{xcolor}
\usepackage{booktabs}
\usepackage{tabularx}
\usepackage{multirow}
\usepackage{comment}

\usepackage{enumitem}
\usepackage{hyperref}
\hypersetup{
    colorlinks=true,
    urlcolor=cyan,
}

\DeclareMathOperator*{\argmax}{argmax}

\def\BibTeX{{\rm B\kern-.05em{\sc i\kern-.025em b}\kern-.08em
    T\kern-.1667em\lower.7ex\hbox{E}\kern-.125emX}}
\begin{document}

\title{Targeting Negative Flips in Active Learning using Validation Sets}

\author{\IEEEauthorblockN{Ryan Benkert, Mohit Prabhushankar, Ghassan AlRegib}
\IEEEauthorblockA{\textit{ OLIVES at the Center for Signal and Information Processing CSIP,} \\
\textit{School of Electrical and Computer Engineering, Georgia Institute of Technology, Atlanta, GA, USA}\\
\{rbenkert3,mohit.p,alregib\}@gatech.edu}
}

\onecolumn 
\begin{description}[leftmargin=2cm,style=multiline]

\item[\textbf{Citation}]{R. Benkert, M. Prabhushankar, and G. AlRegib, "Targeting Negative Flips in Active Learning using Validation Sets", In IEEE International Conference on Big Data 2024, Washington DC, USA.}

\item[\textbf{Review}]{Date of acceptance: 26 October 2024}

\item[\textbf{Code}]{\url{https://github.com/olivesgatech/RoSE}}

\item[\textbf{Bib}] {@ARTICLE\{Benkert2024Targeting,\\ 
author=\{R. Benkert and M. Prabhushankar and G. AlRegib\},\\ 
journal=\{IEEE International Conference on Big Data\},\\ 
title=\{Targeting Negative Flips in Active Learning using Validation Sets\}, \\ 
year=\{2024\},\\ 
month=\{December\},\}
} 

\item[\textbf{Copyright}]{\textcopyright 2024 IEEE. Personal use of this material is permitted. Permission from IEEE must be obtained for all other uses, in any current or future media, including reprinting/republishing this material for advertising or promotional purposes,
creating new collective works, for resale or redistribution to servers or lists, or reuse of any copyrighted component
of this work in other works. }

\item[\textbf{Contact}]{\href{mailto:olives.gatech@gmail.com}{olives.gatech@gmail.com} \\ \url{https://alregib.ece.gatech.edu/} \\ }
\end{description}

\thispagestyle{empty}
\newpage
\clearpage
\setcounter{page}{1}

\twocolumn

\maketitle

\begin{abstract}
The performance of active learning algorithms can be improved in two ways. The often used and intuitive way is by reducing the overall error rate within the test set. The second way is to ensure that correct predictions are not forgotten when the training set is increased in between rounds. The former is measured by the accuracy of the model and the latter is captured in negative flips between rounds. Negative flips are samples that are correctly predicted when trained with the previous/smaller dataset and incorrectly predicted after additional samples are labeled. In this paper, we discuss improving the performance of active learning algorithms both in terms of prediction accuracy and negative flips. The first observation we make in this paper is that negative flips and overall error rates are decoupled and reducing one does not necessarily imply that the other is reduced. Our observation is important as current active learning algorithms do not consider negative flips directly and implicitly assume the opposite. The second observation is that performing targeted active learning on subsets of the unlabeled pool has a significant impact on the behavior of the active learning algorithm and influences both negative flips and prediction accuracy. We then develop \texttt{ROSE} - a plug-in algorithm that utilizes a small labeled validation set to restrict arbitrary active learning acquisition functions to negative flips within the unlabeled pool. We show that integrating a validation set results in a significant performance boost in terms of accuracy, negative flip rate reduction, or both.
\end{abstract}


\section{Introduction}
After machine learning models are deployed for everyday use, \emph{software regression} describes performance deterioration when a model update \cite{regression2021} is performed\footnote{Please note that in this paper, we will be using the term regression to mean \emph{software regression} and not the task of \emph{statistical regression} which predicts continuous values given independent variables.}. As the adoption of deep learning algorithms expands, the analysis of software regression is critical as any potential performance deterioration can have a significant impact on the success of the model \cite{Temel2017_NIPSW}. A field that remains especially prone to software regression is active learning. Active learning \cite{cohn1996active} is a machine learning paradigm that aims to select informative samples that maximize performance while maintaining a limited label budget on the training data. Due to its practicality, active learning has gained momentum within the machine learning community and has been successfully adopted across various industrial scenarios including manufacturing \cite{tong2001active}, robotics \cite{alrobotics}, autonomous vehicles \cite{haussmann2020scalable, kokilepersaud2023focal}, and medical imaging \cite{hoi2006batch, logan2022patient}. Originally, active learning was used to train models before deployment. Recently, however, in applications where data is progressively obtained, new algorithms of deployable active learning have been proposed in subsurface analysis~\cite{benkert2024effective} and medical clinical trial applications~\cite{fowler2023clinical, logan2022decal}.

Software regression occurs in active learning due its iterative nature. At each round, the acquisition function selects a new set of unlabeled samples to annotate and add to the training set. Regression occurs when the model trained on the larger training set predicts samples incorrectly, that the previous model correctly predicted. These samples are called \emph{negative flips} \cite{regression2021}. In spite of the high relevance, active learning algorithms do not optimize for regression and implicitly assume that negative flips reduce when accuracy increases. In Section~\ref{sec:analysis}, we analyze this assumption in detail and find that negative flips can increase \emph{or} decrease, even when the accuracy consistently increases. This contradicts the assumption and motivates integrating software regression into the existing active learning objective.

\begin{figure*}[t]
    \begin{center}
        \includegraphics[scale=0.6]{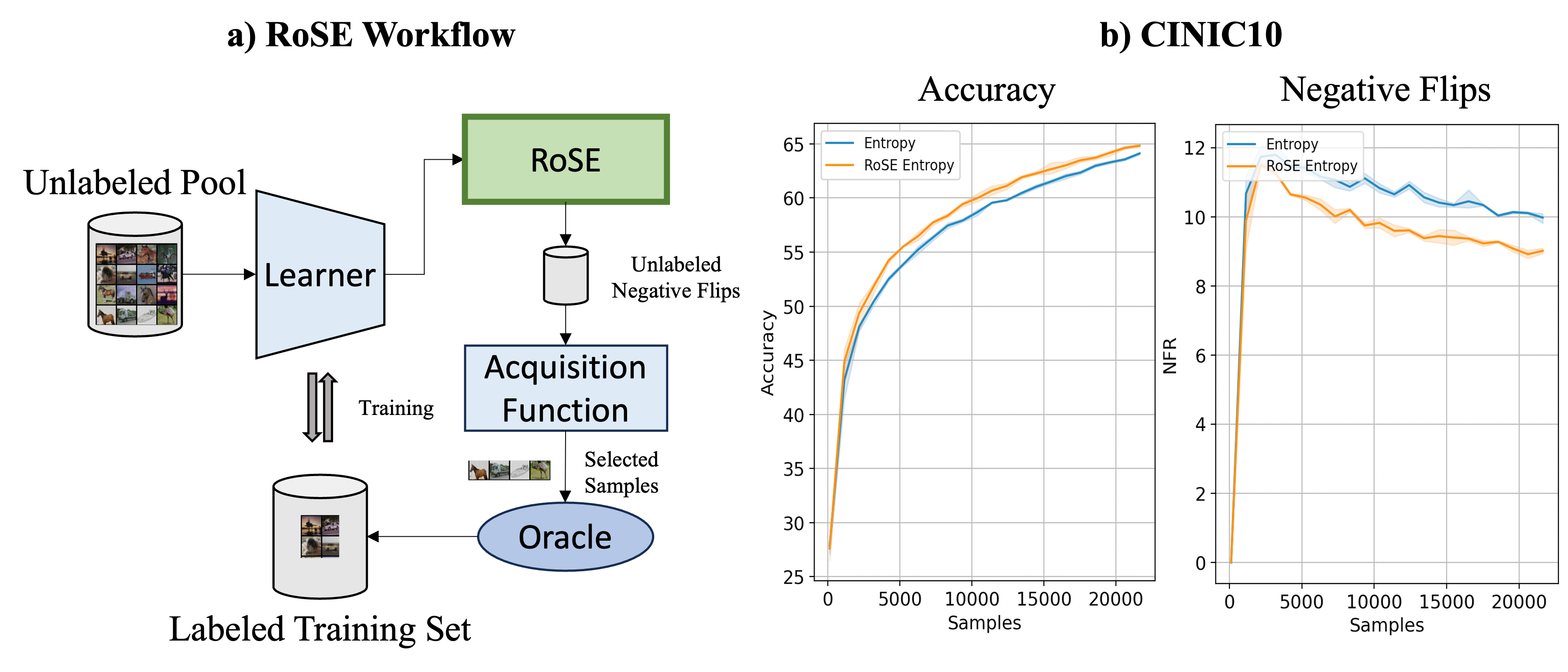}
    \end{center}
    \caption{Overview of our method, \texttt{RoSE}, and learning curves of Entropy sampling on CINIC10. \textbf{a)} high-level diagramm of \texttt{RoSE}. \textbf{b)} accuracy and negative flip rate for the CINIC10 dataset. Higher accuracy and lower negative flip rate values are better.}
    \label{fig:intro}
\end{figure*}

In this paper, we provide a regression-targeted alternative to current active learning frameworks by utilizing small labeled validation sets. In machine learning, validations sets are used for model selection. In active learning, validation sets are popular as they are used for early stopping \cite{morgan1989generalization} to avoid overfitting or for improving acquisition functions in out-of-distribution settings \cite{oodal}. We extend its usage to software regression. We provide a high-level overview of our method in Figure~\ref{fig:intro}a. In each active learning round, we estimate negative flips within the unlabeled pool and restrict the acquisition function to operate on the unlabeled negative flip subset. In Section~\ref{sec:nfr-subsets}, we provide empirical evidence that restricting to negative flips can effectively reduce regression while maintaining or improving accuracy. Our approach, \texttt{RoSE}, is modular and represents a plug-in algorithm that can be deployed with arbitrary active learning algorithms that operate on acquisition functions. In Figure~\ref{fig:intro}b, we show the performance of RoSE in terms of accuracy and negative flips on CINIC10 and note that our algorithm can effectively increase accuracy and/or reduce the number of negative flips. In Section~\ref{sec:results}, we show that \texttt{RoSE} overwhelmingly matches or improves accuracy and/or negative flip performance across a wide variety of datasets, architectures, and acquisition functions.

The contributions of this paper are as follows:
\begin{enumerate}
    \item We provide a comprehensive empirical analysis of regression in active learning which forms the basis of our approach. 
    \item We develop \texttt{RoSE}, a plug-in algorithm to improve negative flips and accuracy simultaneously.
    \item We perform extensive empirical evaluation of \texttt{RoSE} with three datasets, seven acquisition functions, and two architectures.
\end{enumerate}


\section{Background and Prior Art}
\subsection{Active Learning}
Our work relates strongly to existing literature in active learning \cite{dasgupta2011two, settles2009active, hanneke2014theory} and the study of acquisition functions. While the proposed algorithms in literature can effectively reduce the overall test error, they do not optimize for regression or consider regression within the acquisition function. Within this context, the first family of acquisition functions target uncertain samples \cite{wang2014new, benkert2022false, roth2006margin, schohn2000less, benkert2024transitional}. As an example, several approaches define sample importance using softmax probabilities \cite{wang2014new, roth2006margin} where information content is related to the output logits of the network. Closely related works \cite{schohn2000less, tong2001support} select samples based on decision boundary proximity, \cite{Beluch_2018_CVPR} query based on ensemble classification, and \cite{houlsby2011bayesian} query samples based on mutual information. The second family of acquisition functions target samples based on representation within the dataset. A popular group of approaches focus on constructing the core-set of the unlabeled data pool \cite{sener2017active, longtailcoreset}. \cite{gissin2019discriminative} use a discriminative approach, where the authors reformulate the active learning problem as an adversarial training problem. The third family combines uncertain and representative samples within the acquisition function. In several cases \cite{ash2019deep, haussmann2020scalable, benkert2022gauss}, the ranking is established based on generalization difficulty and the representative component is added through interactive sampling. Other examples of this category extend popular bayesian approaches. Here, notable ones are balanced entropy \cite{woo2021active}, batchbald \cite{batchbald}, and power bald \cite{powerbald}. In practice, all three families are frequently used with validation sets to improve the active learning performance. A common setting is early-stopping \cite{morgan1989generalization} where the validation set is used to avoid overfitting. Recent methods such as batchbald \cite{batchbald} explicitly mention early-stopping in their experiments. Other approaches utilize a validation set to improve active learning algorithms alongside other performance settings such as out-of-distribution settings \cite{oodal}. Our work is complementary to these approaches as it provides a further use-case for validation sets in active learning, namely software regression.

\subsection{Software Regression}
In essence, the field of regression is related to the study of continual learning \cite{chen2018lifelong, kirkpatrick2016overcoming, ritter2018online, toneva2018empirical}, incremental learning \cite{prabhu2020gdumb, learningwithoutforgetting2017}, and open set recognition \cite{bendale2016towards, schreierosr2012, lee2021open, kwon2020backpropagated}. Here, catastrophic forgetting of neural networks \cite{kirkpatrick2016overcoming, ritter2018online, toneva2018empirical} represents a particularly relevant topic as ``forgetting" represents the performance deterioration in between different machine learning tasks. A significant research effort involves reducing forgetting through model constraints \cite{aljundi2018memory, kirkpatrick2016overcoming}, loss exploration \cite{shi2021overcoming}, or augmentation \cite{benkert2021explainable, benkert2021explaining, benkert2022example}. While these approaches have strong ties to regression, they focus on reducing forgetting between different tasks, not performance deterioration through updates. Most significantly, our work shares a strong connection with positive congruent training \cite{regression2021} - a process that reduces regression within a single training run. Specifically, the approach borrows concepts from knowledge distillation \cite{hinton2015distilling} and forces representation similarity between the old/updated models through an additional loss term. While the training paradigm effectively reduces regression in single training tasks, it represents a model optimization, not a data selection approach and is therefore not applicable to active learning acquisition functions.

\section{Background}
In this section, we introduce definitions and notations for both active learning and regression. We start with the problem setting of active learning and further introduce a metric for measuring regression.

\subsection{Active Learning}
The objective of active learning \cite{cohn1996active} involves maximizing performance under minimal data annotations. We consider a dataset $D$ where the subset $D_{train}$ represents the current set of annotated labeled examples and $D_{pool}$ the unlabeled target subset from which we select the next set of samples. In batch active learning, the most informative batch of $b$ samples $X^* = \{x^*_1, ..., x^*_b\}$ are selected for a predefined target task. Formally, we write

\begin{equation}
\label{eq:aquisition-function}
    X^* = \argmax_{x_1, ..., x_b \in D_{pool}} a(x_1, ..., x_b | h_w(x)).
\end{equation}
where $a$ represents the acquisition function and $h_w$ the deep model conditioned on the parameters $w$. In current active learning algorithms, $a$ is optimized to reduce the overall error and performance regression is not considered. 


\subsection{Negative Flip Rate}
To measure regression in active learning, we consider a non-linear discriminative model $h^{old}_w$ and an updated version, $h^{new}_w$, where $h^{new}_w$ is trained on more data than $h^{old}_w$. The prediction $\Tilde{y_i}$ of the current model $h^{new}_w$ belongs to one of four categories: 1) predicted correctly by both models (both correct), 2) predicted incorrectly by both models (both wrong), 3) predicted incorrectly by $h^{old}_w$ and correctly by $h^{new}_w$ (positive flips), and 4) predicted correctly by $h^{old}_w$ and incorrectly by $h^{new}_w$ (negative flips). In this paper, we measure regression with the negative flip rate\cite{regression2021}, given by:

\begin{equation}
\label{eq:softmax-distribution}
    NFR = \frac{1}{N}\sum_{i = 1}^{N} \mathbf{1}_{\tilde{y}^{old}_i = y_i, \tilde{y}^{new}_i != y_i},
\end{equation}
where $\mathbf{1}_{\tilde{y}^{old}_i = y_i, \tilde{y}^{new}_i != y_i}$ represents a binary variable that reduces to one in the case of a negative flip or zero otherwise. The ratio is called negative flip rate or NFR in short.

\section{Empirical Analysis}
\label{sec:analysis}
\subsection{Regression in Active Learning}
\begin{figure*}[!ht]
    \begin{center}
        \includegraphics[scale=0.73]{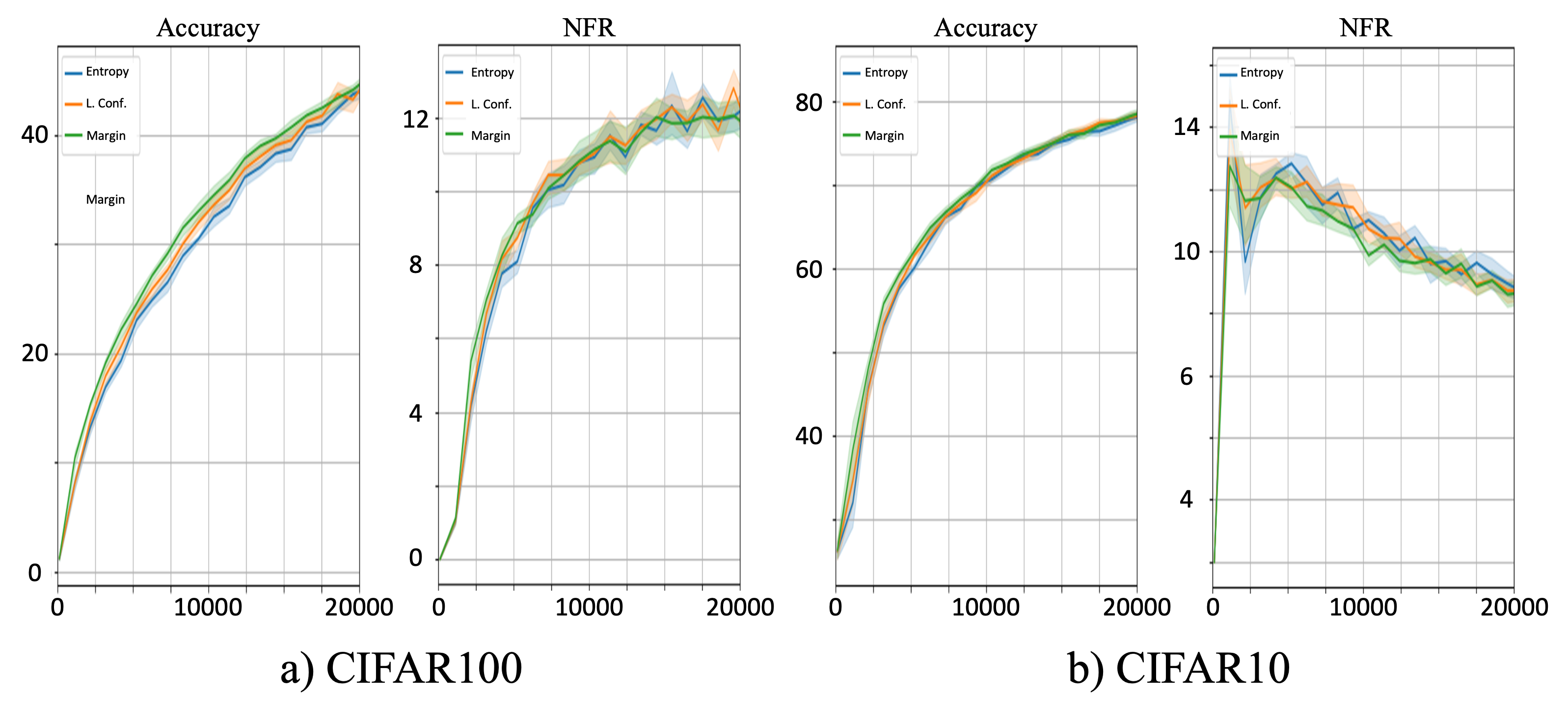}
    \end{center}
    \caption{Accuracy and NFR curves of three different acquisition functions on the CIFAR10 and CIFAR100 benchmark. The plots consider entropy sampling, margin sampling, and least confidence sampling. For accuracy, higher values indicates strong performance. For NFR, lower values are preferred. \textbf{a)} CIFAR100 accuracy and NFR performance. \textbf{b)} CIFAR10 accuracy and NFR performance.}
    \label{fig:nfr-analysis}
\end{figure*}

In this paper, we integrate software regression reduction into the active learning objective. To this end, we start by investigating how software regression manifests within active learning frameworks and to what extent existing algorithms address it. For this purpose, we first compare the performance of several popular active learning algorithms in terms of software regression, and overall error rate reduction. We measure software regression via NFR where a low NFR represents less software regression. The overall error rate is measured via classification accuracy where a higher accuracy value indicates a lower overall error rate. In our analysis we perform active learning with three acquisition functions: entropy sampling \cite{wang2014new}, least confidence sampling \cite{wang2014new}, and margin sampling \cite{roth2006margin}. We show their performance in Figure~\ref{fig:nfr-analysis} where the x-axis represents the labeled training set size and the y-axis shows the accuracy or NFR respectively. \emph{While all acquisition functions increase the accuracy in between rounds, the NFR is either increased or decreased with additional samples.} Specifically, NFR decreases on CIFAR10 and increases on the more complex CIFAR100 dataset. This observation is important as it shows that optimizing for accuracy alone does not necessarily lead to a reduction in regression and motivates the need of explicitly optimizing for regression in active learning. In the following paragraphs, we delve deeper into several sources of regression. Specifically, we consider class complexity and class imbalance.



\begin{figure}[!ht]
    \begin{center}
        \includegraphics[scale=0.32]{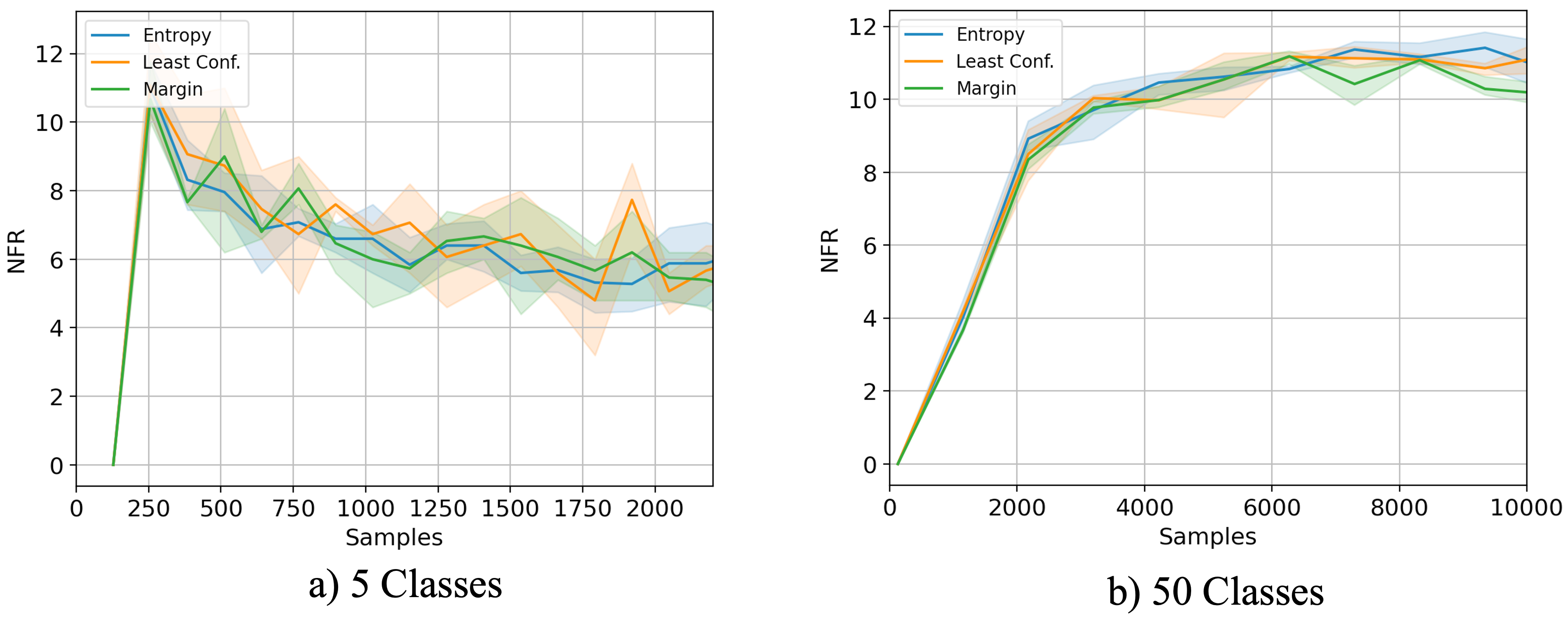}
    \end{center}
    \caption{Class complexity analysis of the NFR. Each plot represents a different number of classes from CIFAR100. \textbf{a)} 5 classes; \textbf{b)} 50 classes.}
    \label{fig:nfr-class-complexity}
\end{figure}

\paragraph{Class Complexity}
In Figure~\ref{fig:nfr-class-complexity}, we investigate the effect of class complexity on the NFR. For this purpose, we perform active learning with subsets of CIFAR100 containing 5, 20, 50, or 100 of the original 100 classes within the training and test sets. Within Figure~\ref{fig:nfr-class-complexity}, we show the learning curves for 5 classes and 50 classes. Our experimental setup is equivalent to our previous experiments with the exception of using random crop and random horizontal flips as additional training augmentations. Further, we use a query size of 128 on the subset containing 5 classes due to the small size of the unlabeled data pool (2500 of the original 50000 samples). Overall, we note that class diversity significantly impacts the NFR trend. Settings containing fewer classes (5 classes or 20 classes) decrease in NFR with additional samples while more complex settings (50 or 100 classes) increase in NFR when samples are added to the training pool. We reason that settings with fewer classes establish robust class representations earlier due to the larger amount of samples per class. In contrast, higher class complexities result in fewer samples per class and additional data points can cause more significant shifts to the decision boundary. 

\begin{figure}[!ht]
    \begin{center}
        \includegraphics[scale=0.32]{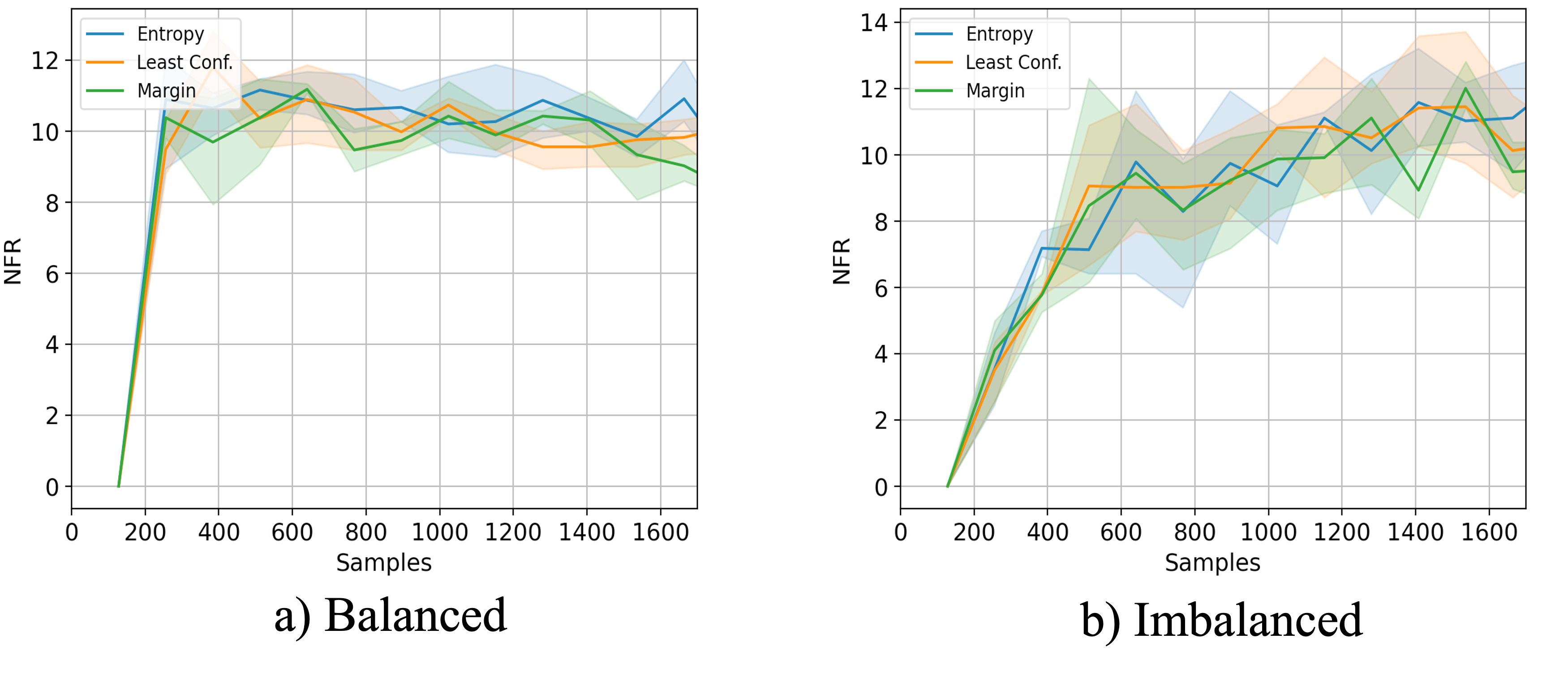}
    \end{center}
    \caption{NFR on 15 classes in the CIFAR100 dataset. \textbf{a)} balanced class setting for both unlabeled pool and test set. \textbf{b)} imbalanced setting.}
    \label{fig:nfr-imbalance}
\end{figure}

\paragraph{Class Imbalance}
We analyze the impact of class imbalance on the NFR in Figure~\ref{fig:nfr-imbalance}. For our experiments, we select a subset of 15 classes from the CIFAR100 dataset and introduce imbalance within the training and test set: we reduce the number of training samples for classes 1-7 from 500 data points to 100 points and the test samples of classes 8-15 from 100 samples to 10. As a result, the training set contains underrepresented classes (index 8-15) that are well represented within the test set. Further, classes 8-15 are over-represented in the training set but underrepresented in the test distribution. Similar to our previous analysis, class imbalance significantly impacts the NFR trend. The NFR decreases in the balanced setting but increases when imbalance is introduced. We reason that the distribution shift caused by imbalance causes more decision boundary fluctuations on the test set. When samples within the training set are underrepresented and rare, the model is more prone to overfitting to the sparse training set resulting in more negative flips in the subsequent round.

\subsection{Regression Reducing Subsets}
\label{sec:nfr-subsets}
\begin{figure}
    \begin{center}
        \includegraphics[scale=0.32]{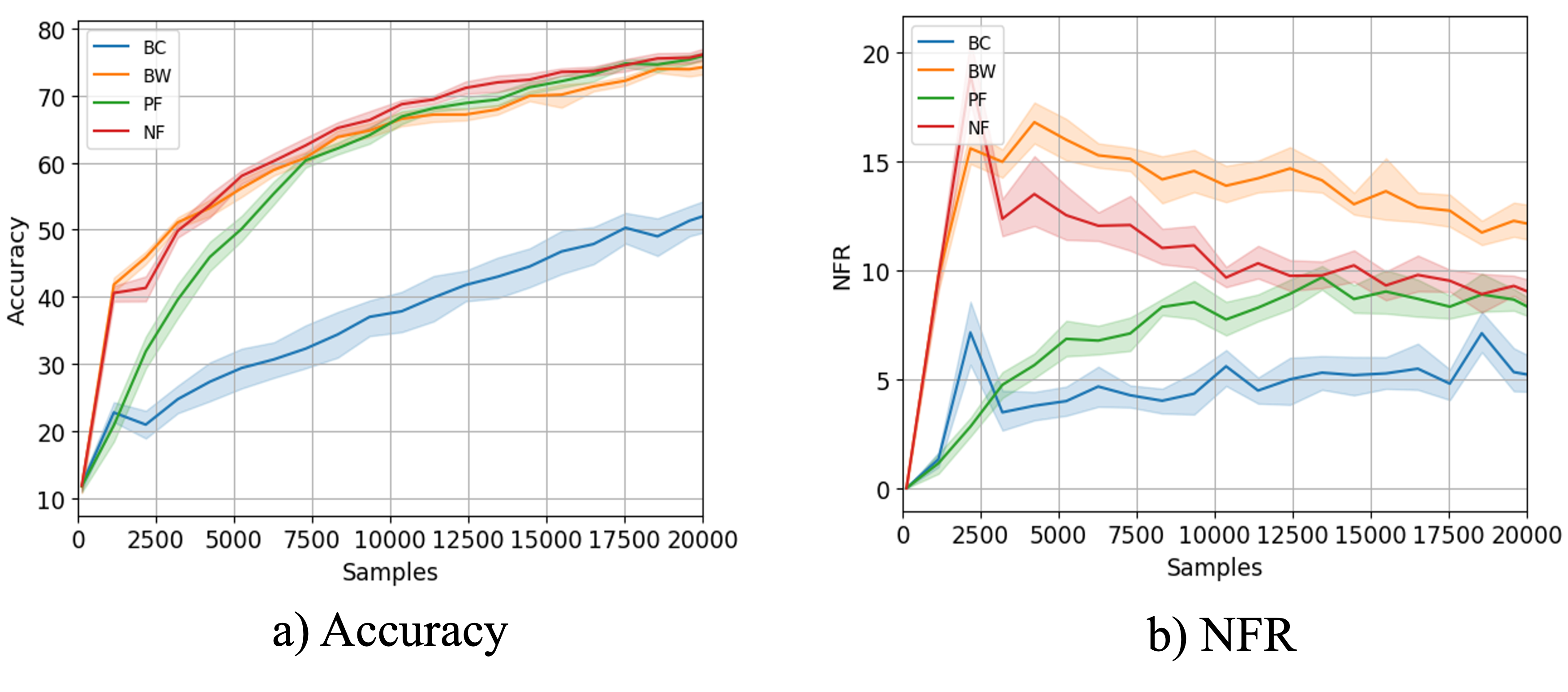}
    \end{center}
    \caption{Random sampling from the restricted unlabeled pool based on the subsets both correct (BC), both wrong (BW), negative flips (NF), and positive flips (PF). \textbf{a)} Accuracy; \textbf{b)} NFR.}
    \label{fig:subset-performance}
\end{figure}

Within this section, we empirically investigate which samples are likely to reduce the NFR on the test set during active learning. Here, a natural set of candidates are the four sample categories caused by software regression: both correct, both wrong, positive flips, and negative flips. For this purpose, we consider an idealised scenario where we have access to the labels within $D_{pool}$ and have perfect knowledge which samples belong to the subsets both correct/wrong as well as negative/positive flips. In each active learning round, we first partition the unlabeled pool to the four subsets, and limit our selection pool to the respective subset exclusively. In Figure~\ref{fig:subset-performance}, we show the learning curves when we randomly sample from the restricted search pool based on both correct (BC), both wrong (BW), negative flips (NF), and positive flips (PF). Overall, we note that the subset partitioning significantly impacts the accuracy and NFR learning curve even though the same acquisition function (random sampling) is used. While all subsets increase the accuracy of the model, the NFR plot shows opposing trends between the subsets. Specifically, the BC and PF plots have increasing NFR trends while BW and NF have decreasing NFR trends. Furthermore, the both correct subset has the lowest NFR but does not contain informative samples that increase the accuracy significantly, thereby having lowest accuracy. Both wrong contains informative samples that increase the accuracy but that are also ambiguous and result in fluctuating predictions with a higher NFR. Positive flips are similar to both correct subset and exhibits low NFR values at the cost of accuracy in early rounds. Finally, the negative flip subset consistently performs among the highest in terms of accuracy while still reducing the NFR in comparison to both wrong. This is especially surprising as the three subsets both wrong, negative flips, and positive flips perform similar in terms of accuracy in later active learning rounds. Hence, we reason that sampling from negative flips has the potential to improve both accuracy and NFR in active learning algorithms.


\section{Our Method}
\subsection{Regression Targeted Active Learning}
Within the previous section, we observed that negative flips in the unlabeled pool increase accuracy while providing more performance consistency. Hence, negative flip samples have the potential to improve the performance of existing algorithms both in terms of software regression and overall error rate. For this purpose, our algorithm estimates the negative flip subset and applies the acquisition function on top of the smaller subspace in a plug-in fashion. Mathematically, this is equivalent to modifying the active learning objective in Equation~\ref{eq:aquisition-function} with an additional subset estimation step defined by the function $q_{\phi}$ with hyperparameters $\phi$:

\begin{equation}
\label{eq:restricted-acquisition-function}
    X^* = \argmax_{x_1, ..., x_b \in q_{\phi}(D_{pool})} a(x_1, ..., x_b | h^{new}_w(x)).
\end{equation}

Our approach is practical as it does not impose any restrictions on the acquisition function and can be combined with arbitrary active learning algorithms defined on acquisition functions. In the following subsections, we describe our algorithm as an instance of $q_{\phi}$ that approximates the negative flip subset.

\begin{figure*}
    \begin{center}
        \includegraphics[scale=0.55]{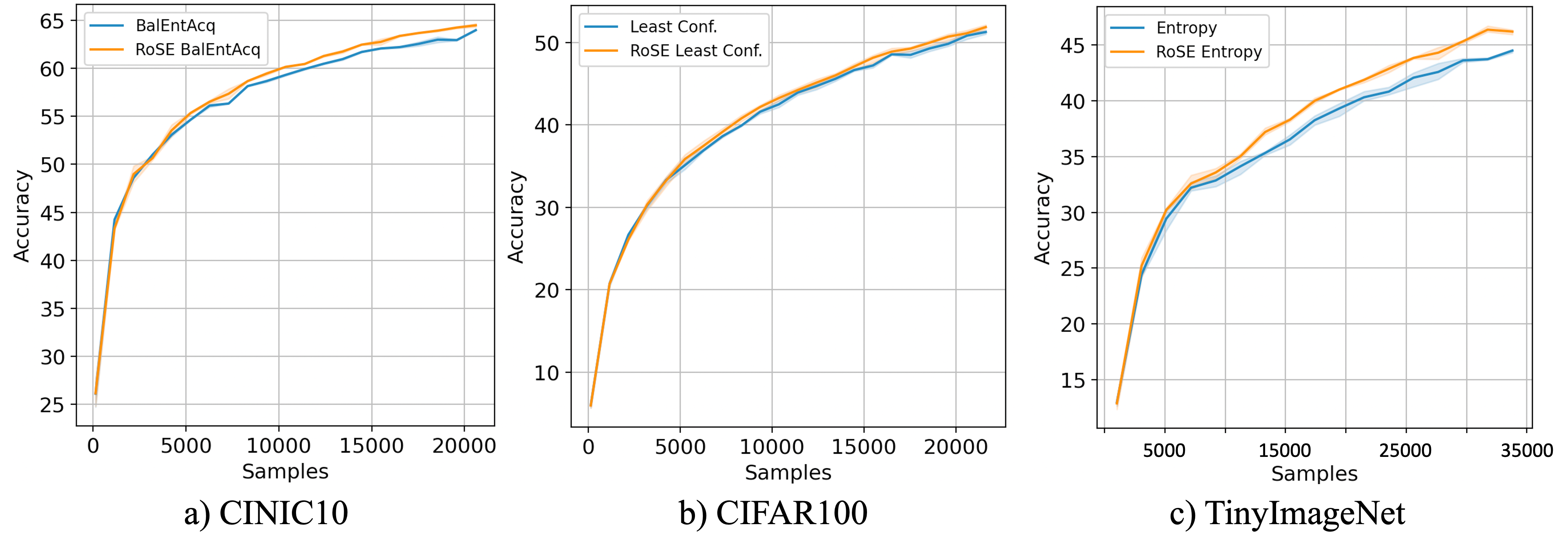}
    \end{center}
    \caption{Accuracy learning curves on three dataset benchmarks with different acquisition functions. The x-axis shows the labeled training set size. \textbf{a)} BalEntAcq with and without \texttt{RoSE} on CINIC10. \textbf{b)} Least confidence sampling with and without \texttt{RoSE} on CIFAR100. \textbf{c)} Entropy sampling with and without \texttt{RoSE} on TinyImageNet.}
    \label{fig:acc-diff-strats}
\end{figure*}
\subsection{Algorithm}

Our algorithm consists of two stages: 1) search space reduction; and 2) misprediction estimation. In 1) we partition the unlabeled pool into two union sets $S_{PN}$ and $S_{CW}$ where $S_{PN}$ is the union of both negative and positive flips. In 2) we perform misprediction detection by utilizing a small labeled validation set. Applied to $S_{PN}$, the estimated mispredicted samples constitute the negative flips. We call our method \texttt{Regression-ordered Subset Estimation} or \texttt{RoSE} in short. In the following, we describe \texttt{RoSE} in detail.

\paragraph{Search Space Reduction}
A fundamental observation to reduce the search space is that the union of the positive and negative flips can be derived analytically through prediction switches. Consider the four subsets both correct $S_C$, both wrong $S_W$, positive flips $S_P$, and negative flips $S_N$ that together form the unlabeled pool $D_{pool}=\bigcup_{i \in \{S_C, S_W, S_P, S_N\}}S_i$. The unions $S_C \cup S_W$ and $S_P \cup S_N$ are separable by their prediction switches in between rounds

\begin{equation}
	\label{eq:union-estimation}
	\begin{split}
		&S_{PN} = S_P \cup S_N = \{ x_i \in D_{pool}: s_i = \mathbf{1}_{\Tilde{y_i}^{new} != \Tilde{y_i}^{old}} = 1 \}\\
		&S_{CW} = S_C \cup S_W = \{ x_i \in D_{pool}: s_i = \mathbf{1}_{\Tilde{y_i}^{new} != \Tilde{y_i}^{old}} = 0 \}\\
	\end{split}
\end{equation}

where $s_i \in \{0, 1\}$ indicates whether the model from the previous round predicts a different class than the current model. Equation~\ref{eq:union-estimation} is intuitive. Samples with the same prediction in between rounds are either both correct and both wrong while a prediction switch indicates a flip, regardless if it is positive or negative. 

\paragraph{Misprediction Detection}
Our misprediction detection requires several steps: first, we generate misprediction scores for all samples in $S_{PN}$ using well-established misprediction detection approaches \cite{gal2016dropout, lakshminarayanan2017simple, liu2021energybased}. Second, we determine an appropriate score threshold for classifying negative flips and partition the union set. For the misprediction scores, we utilize the energy-based approach by \cite{liu2021energybased} as it represents a recent algorithm with strong detection performance that can be computed within a single forward pass. This allows a seamless integration into active learning workflows and provided sufficient results in our experiments. Determining an appropriate threshold is more involved as the optimal threshold may fluctuate in different active learning rounds. For this purpose, we set the threshold to reflect the accuracy on a small labeled validation set. Specifically, given a score function $g$ we select $k = (1-acc_{val})*|D_{pool}|$ samples with the highest misprediction score as our negative flip subset

\begin{equation}
	\label{eq:nf-subset-estimation}
	\begin{split}
		\Tilde{S}_{N} = q_{\phi}(D_{pool}) = \argmax_{x_1, ..., x_k \in S_{PN}} \sum_{i}^k g(x_i).
	\end{split}
\end{equation}

Here, the hyperparameter $\phi = acc_{val}$ represents the accuracy on the validation set.
\section{Experiments}
\label{sec:results}
In this section, we provide experimental results to answer the following questions: 1) how well does \texttt{RoSE} perform under several datasets with different types of complexities?; 2) does \texttt{RoSE} perform well with a wide variety of acquisition functions?; and 3) does performance enhancement via \texttt{RoSE} manifest as a net-improvement when considering both regression and overall error rate? 

\begin{figure*}
    \begin{center}
        \includegraphics[scale=0.55]{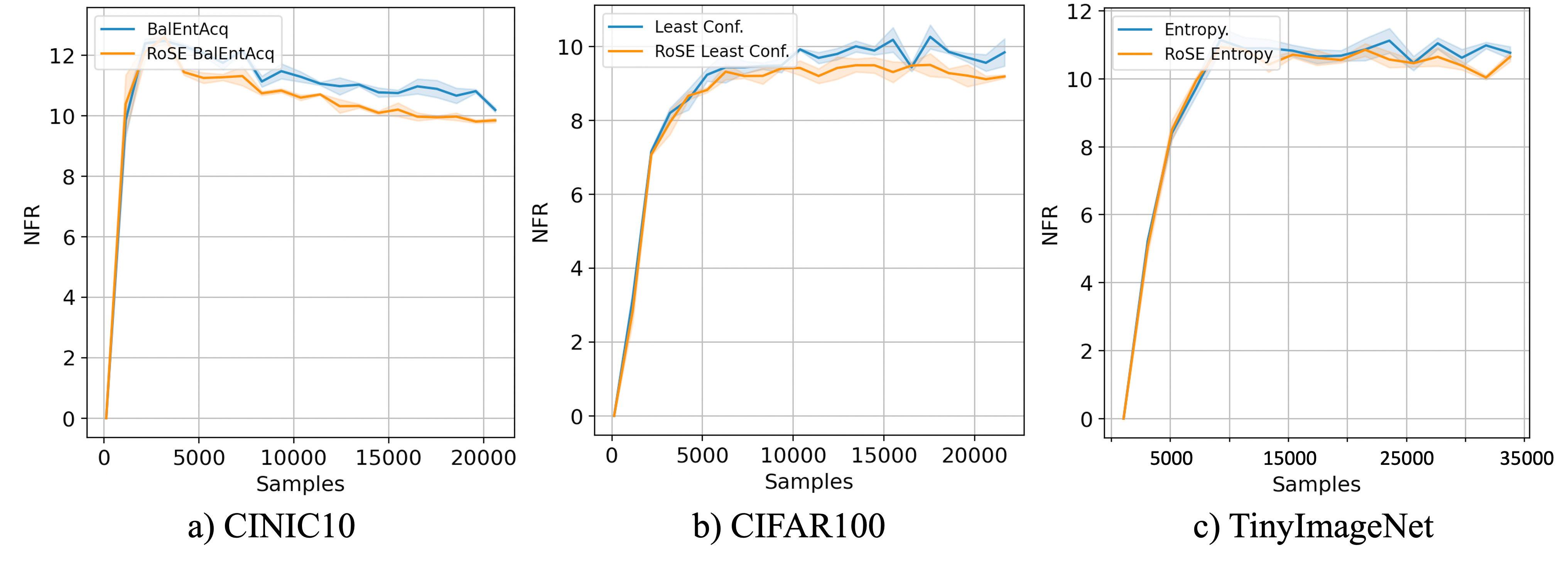}
    \end{center}
    \caption{NFR learning curves on three dataset benchmarks with different acquisition functions. The x-axis shows the labeled training set size. \textbf{a)} BalEntAcq with and without \texttt{RoSE} on CINIC10. \textbf{b)} Least confidence sampling with and without \texttt{RoSE} on CIFAR100. \textbf{c)} Entropy sampling with and without \texttt{RoSE} on TinyImageNet.}
    \label{fig:nfr-diff-strats}
\end{figure*}
\begin{figure*}
    \begin{center}
        \includegraphics[scale=0.55]{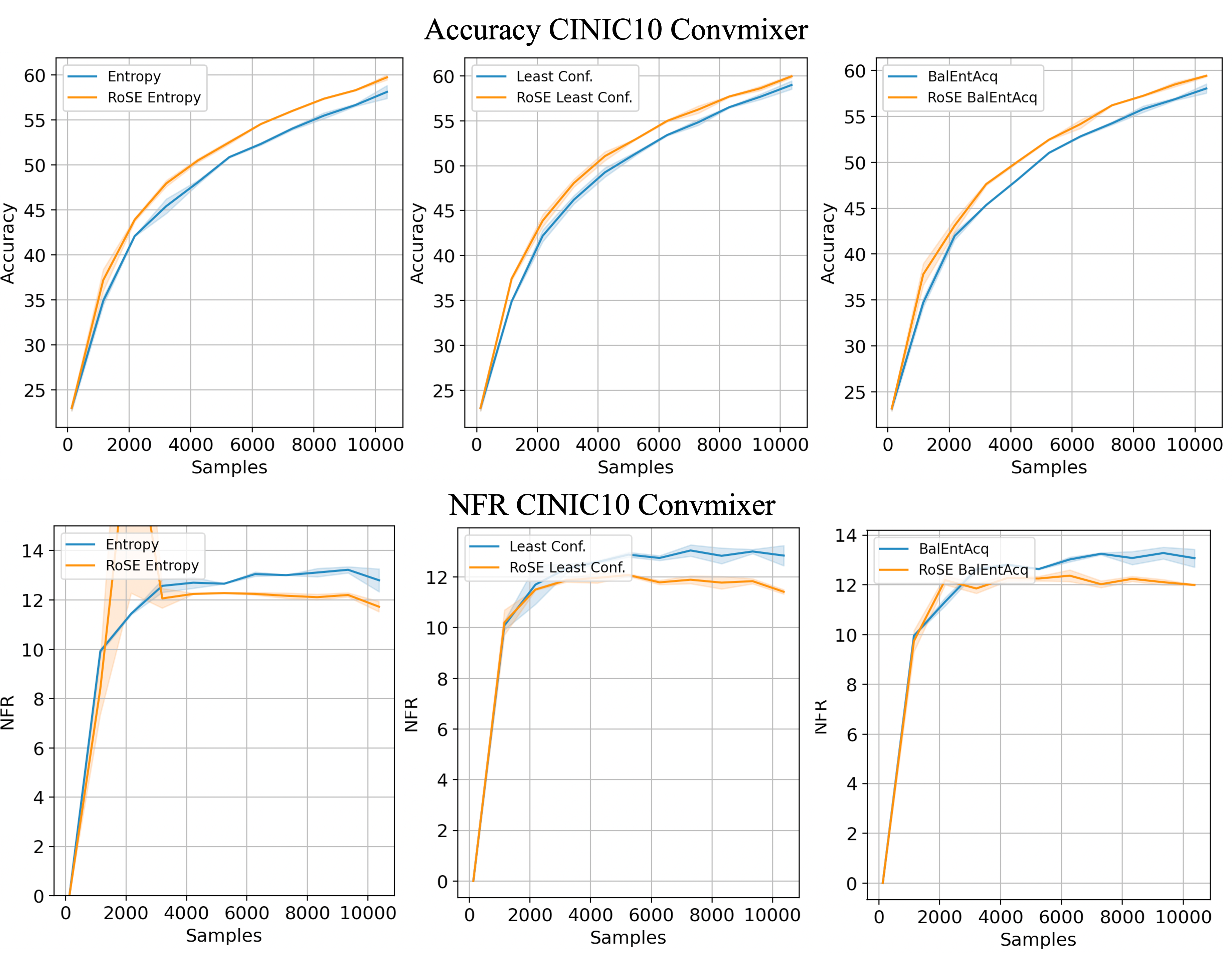}
    \end{center}
    \caption{Convmixer learning curves for CINIC-10 dataset}
    \label{fig:convmix-all}
\end{figure*}
\subsection{Experimental Setup}
We benchmark \texttt{RoSE} by comparing against the acquisition functions directly applied on the entire unlabeled pool without limiting the search space. For this purpose, we consider a wide range of popular acquisition functions, namely Entropy \cite{wang2014new}, Margin \cite{roth2006margin}, Least Confidence \cite{wang2014new}, BADGE \cite{ash2019deep}, BALD \cite{houlsby2011bayesian}, PowerBALD \cite{powerbald}, and BalEntAcq \cite{woo2021active}. Entropy, margin, and least confidence represent popular active learning algorithms that do not require additional changes to the architecture and operate on top of the neural network representation. BADGE also does not require architecture changes but utilizes gradient embeddings instead of the neural network representation. Finally, BALD, PowerBALD, and BalEntAcq are approaches that require modifications to the neural network architecture. In our case, we add a shallow bayesian neural network on top of the feature extractor and perform Monte-Carlo-Dropout during the acquisition phase \cite{gal2017deep}. We consider three dataset benchmarks: CIFAR100 \cite{krizhevsky2009learning}, CINIC10\cite{cinic10}, and TinyImageNet \cite{le2015tiny}. CINIC10 represents an interesting active learning benchmark as it is large (270k data points) and the data originates from two separate sources. Hence, the algorithm must select samples from both sources to achieve strong performance. Both CIFAR100 and TinyImageNet represent benchmarks with a higher class and data complexity. CIFAR100 contains 100 classes with images of size 32x32. TinyImageNet contains 200 classes and has a higher image resolution of 64x64. In our experiments, we consider a resnet-18 \cite{he2016deep} architecture as well as the recent ViT inspired convmixer \cite{trockman2022patches}. For all benchmarks except TinyImageNet, we start with an initial randomly chosen 128 samples and query 1024 additional samples in subsequent active learning round. For TinyImageNet, we start with 1024 random samples and query 2048 every round. Our validation set is based on dataset size and class complexity. For CINIC10, we consider 1\% of the original training set as our labeled validation set. For CIFAR100 and TinyImageNet, we select 10\% and 5\% respectively. For ease of experimentation, we do not implement early stopping but note that adding it during the training phase can further increase the active learning performance \cite{batchbald}. In each round, we optimize with the ADAM optimizer and a learning rate of 0.0001. Further, we train until either the training accuracy reaches 98\% or 200 epochs have passed. We train each model from scratch every round and average our results over multiple random seeds. Due to space limitations, we report a subset of our numerical and graphical results in the paper.

\begin{table*}
\centering
\caption{Percentage where \texttt{RoSE} outperforms the baseline in $Acc$ and $NFR$.}
\label{tab:overall-rn18}
\small
\begin{tabularx}{\textwidth}{X c c c c c c}
\hline
                       & \multicolumn{2}{c}{CINIC10} & \multicolumn{2}{c}{CIFAR100} & \multicolumn{2}{c}{TinyImageNet}  \\
              Algorithm & $\Delta Acc$ & $\Delta NFR$ & $\Delta Acc$ & $\Delta NFR$& $\Delta Acc$ & $\Delta NFR$   \\
		\hline
		\hline
		Entropy \cite{wang2014new}      & 0.924±0.043      & 0.924±0.021      & 0.924±0.043      & 0.697±0.077      & 0.922±0.028      & 0.588±0.127        \\
		Margin \cite{roth2006margin}    & 0.288±0.204      & 0.894±0.043      & 0.545±0.129      & 0.591±0.129      & 0.794±0.029      & 0.647±0.059    \\
		Least Conf. \cite{wang2014new}  & 0.939±0.021      & 0.955±0.000      & 0.803±0.119      & 0.758±0.057      & 0.971±0.029      & 0.735±0.029    \\
		BADGE \cite{ash2019deep}        & 0.803±0.130      & 0.697±0.077      & 0.591±0.000      & 0.432±0.023      & 0.550±0.150      & 0.450±0.050    \\
		BalEntAcq \cite{woo2021active}  & 0.881±0.024      & 0.857±0.000      & 0.977±0.023      & 0.750±0.068      & 0.912±0.088      & 0.206±0.029    \\
		BALD \cite{houlsby2011bayesian} & 0.500±0.024      & 0.571±0.048      & 0.682±0.091      & 0.591±0.045      & 0.765±0.118      & 0.500±0.088    \\
		PowerBALD \cite{powerbald}      & 0.786±0.024      & 0.429±0.095      & 0.795±0.068      & 0.500±0.091      & 0.882±0.000      & 0.382±0.029    \\
		\hline
\end{tabularx}

\end{table*}
\subsection{Learning Curves}
We start by analyzing the learning curves of \texttt{RoSE} on the three dataset benchmarks using resnet-18. In Figure~\ref{fig:acc-diff-strats} and \ref{fig:nfr-diff-strats}, we report the performance in terms of accuracy and NFR respectively. In all plots, the x-axis represents the training set size. Due to space constraints, we show results on three acquisition functions, but similar observations are made across other methods. Overall, we observe that \texttt{RoSE} matches or improves the acquisition function performance across all three benchmarks. In particular, we note a significant accuracy improvement when using \texttt{RoSE} over the baseline (Figure~\ref{fig:acc-diff-strats}). For instance, using \texttt{RoSE} on top of entropy sampling results in a 2\% improvement on the TinyImageNet benchmark in the final active learning round. We reason that limiting the unlabeled pool to negative flips increases information content in the training set by rejecting uninformative samples that would have otherwise been added to the acquisition batch. From Figure~\ref{fig:nfr-diff-strats}, we further observe that the accuracy improvement correlates with an NFR decrease when \texttt{RoSE} is used in combination with the acquisition function. This indicates that the combination of restricting the unlabeled pool to negative flips can help in rendering the network more robust to regression. In addition, we observe that \texttt{RoSE} is particularly effective in later rounds and tends to match the baseline performance in earlier rounds. We reason that \texttt{RoSE} is more accurate in estimating negative flips when more labeled training data is available and the representation space is less prone to mapping positive/negative flips to similar misprediction scores.

We further provide learning curves using the convmixer architecture in Figure~\ref{fig:convmix-all} on CINIC-10 dataset. We tested the convmixer architecture with the acquisition functions Entropy \cite{wang2014new}, Least Confidence sampling \cite{wang2014new} and BalEntAcq \cite{woo2021active}.

\subsection{Numerical Analysis}
\paragraph{Overall Performance}
We evaluate the overall performance of \texttt{RoSE} in Table~\ref{tab:overall-rn18}. For this purpose, we calculate the ratio of rounds in which \texttt{RoSE} outperforms the baseline in either accuracy or NFR. Formally, given the number of total rounds $N_{total}$ as well as the number of rounds in which \texttt{RoSE} outperforms the baseline $N_{\texttt{RoSE}}$, we calculate the ratio as $\frac{N_{total}}{N_{\texttt{RoSE}}}$. A high value larger than 0.5 indicates that \texttt{RoSE} outperforms in the majority of rounds while a value smaller than 0.5 indicates the opposite. As most values are larger than 0.5, we note that our method matches or outperforms the baseline in the majority of cases both in NFR and accuracy. In particular, we note that \texttt{RoSE} frequently outperforms both in NFR and accuracy. For instance, least confidence sampling exhibits high ratio values both for NFR and accuracy across all three datasets. This indicates that negative flips act complementary to each other by adding more information to the training pool while further increasing consistency. We further acknowledge the cases in which \texttt{RoSE} results in smaller ratio than 0.5. This is true for the accuracy ratio for margin sampling on CINIC10 as well as the NFR ratio for PowerBALD on TinyImageNet. While the performance for one metric is low, we note that the other exhibits a strong ratio. Our observation indicates that while \texttt{RoSE} is beneficial in most cases, sampling from negative flips can also act in opposition to either NFR or accuracy. 

\paragraph{Selected Rounds}
In addition to learning curves, we further analyze the performance of \texttt{RoSE} in individual active learning rounds. In Table~\ref{tab:cinic10-rn18}, we show both accuracy and NFR for several selected active learning rounds for all algorithms on the CINIC10 benchmark. Specifically, we show the performance in round 5, 10, 15, and 20. Similar to our previous analysis, we observe that \texttt{RoSE} matches or outperforms the baseline acquisition function across a wide range of acquisition functions. In particular, we note that in the few cases where \texttt{RoSE} does reduce the accuracy of the model, the NFR is reduced as well. This can be seen in round 10 and 15 when margin sampling is performed. Here, both accuracy and NFR are reduced in combination with \texttt{RoSE} indicating favorable performance regression properties of our approach. 

\begin{table*}
\centering
\caption{$Acc$ and $NFR$ in \% for several rounds on CINIC10.}
\label{tab:cinic10-rn18}
\begin{tabularx}{0.86\textwidth}{l c|c|c|c|c|c|c|c|c}
\hline
            $D_{train}/D_{pool}$ &           & \multicolumn{2}{c}{5248/89100} & \multicolumn{2}{c}{10368/89100} & \multicolumn{2}{c}{15488/89100}  & \multicolumn{2}{c}{20608/89100}\\
              Algorithm & $q_{\phi}(D_{pool})$ & $Acc$ & $NFR$ & $Acc$ & $NFR$& $Acc$ & $NFR$& $Acc$ & $NFR$   \\
		\hline
		\hline
        \multirow{2}{*}{Entropy \cite{wang2014new}}
		& Baseline & 53.8±0.1      & 11.4±0.1      & 58.6±0.2      & 10.8±0.1     & 61.5±0.1      & 10.3±0.1     & 63.5±0.1      & 10.1±0.0      \\
		& Ours     & \bf{55.5±0.0} & \bf{10.5±0.1} & \bf{60.0±0.3} & \bf{9.8±0.1} & \bf{62.6±0.4} & \bf{9.3±0.2} & \bf{64.6±0.1} & \bf{8.9±0.1} \\
        \hline
        \multirow{2}{*}{Margin \cite{roth2006margin}}
		& Baseline & 55.5±0.2 & 10.6±0.4 & \bf{60.1±0.1} & 10.1±0.1     & \bf{62.6±0.0} & 9.7±0.1      & 64.6±0.2 & 9.4±0.1      \\
		& Ours     & 55.5±0.4 & 10.1±0.2 & 59.8±0.1      & \bf{9.5±0.1} & 62.3±0.1     & \bf{9.3±0.1} & 64.3±0.1 & \bf{8.8±0.1} \\
        \hline
        \multirow{2}{*}{Least Conf. \cite{wang2014new}}
		& Baseline & 54.4±0.1      & 11.3±0.3      & 59.2±0.2      & 10.7±0.2      & 61.9±0.1      & 10.3±0.2      & 64.0±0.1     & 9.9±0.0      \\
		& Ours     & \bf{55.6±0.0} & \bf{10.4±0.2} & \bf{59.9±0.2} & \bf{9.4±0.2} & \bf{62.5±0.2} & \bf{9.2±0.0} & \bf{64.3±0.1} & \bf{8.9±0.0} \\
        \hline
        \multirow{2}{*}{BADGE \cite{ash2019deep}}
		& Baseline & 55.6±0.3 & 10.5±0.1 & 60.1±0.2 & 10.0±0.2 & 62.7±0.1      & 9.6±0.2      & 64.5±0.1      & 9.4±0.1      \\
		& Ours     & 55.9±0.4 & 10.5±0.2 & 60.4±0.5 & 9.8±0.1  & \bf{63.0±0.1} & 9.4±0.1 & \bf{64.9±0.1} & \bf{9.1±0.0} \\
        \hline
        \multirow{2}{*}{BalEntAcq \cite{woo2021active}}
		& Baseline & 54.7±0.1      & 12.1±0.1      & 59.3±0.1      & 11.3±0.2      & 62.1±0.0      & 10.8±0.1      & 64.0±0.1      & 10.2±0.1      \\
		& Ours     & \bf{55.3±0.1} & \bf{11.3±0.2} & \bf{60.2±0.1} & \bf{10.6±0.1} & \bf{62.8±0.3} & \bf{10.2±0.2} & \bf{64.5±0.2} & \bf{9.8±0.1} \\
        \hline
        \multirow{2}{*}{BALD \cite{houlsby2011bayesian}}
		& Baseline & 54.3±0.0      & 12.0±0.2      & 59.9±0.1 & 10.8±0.1      & 62.7±0.2      & 10.3±0.0      & 64.6±0.1      & 10.0±0.1      \\
		& Ours     & \bf{54.9±0.1} & \bf{11.6±0.2} & 60.0±0.1 & 10.6±0.2      & 62.6±0.0      & \bf{10.2±0.0} & 64.5±0.0 & \bf{9.8±0.1} \\
        \hline
        \multirow{2}{*}{PowerBALD \cite{powerbald}}
		& Baseline & 54.7±0.4      & 11.7±0.5      & 59.8±0.5      & \bf{10.6±0.2} & 62.2±0.1      & 10.3±0.1      & 63.9±0.2      & 9.9±0.2      \\
		& Ours     & 55.3±0.5      & 11.4±0.1      & 59.7±0.2      & 11.0±0.1      & \bf{62.4±0.0} & 10.5±0.0 & \bf{64.6±0.0} & 9.8±0.0 \\
        \hline
\end{tabularx}

\end{table*}
\begin{figure*}
    \begin{center}
        \includegraphics[scale=0.55]{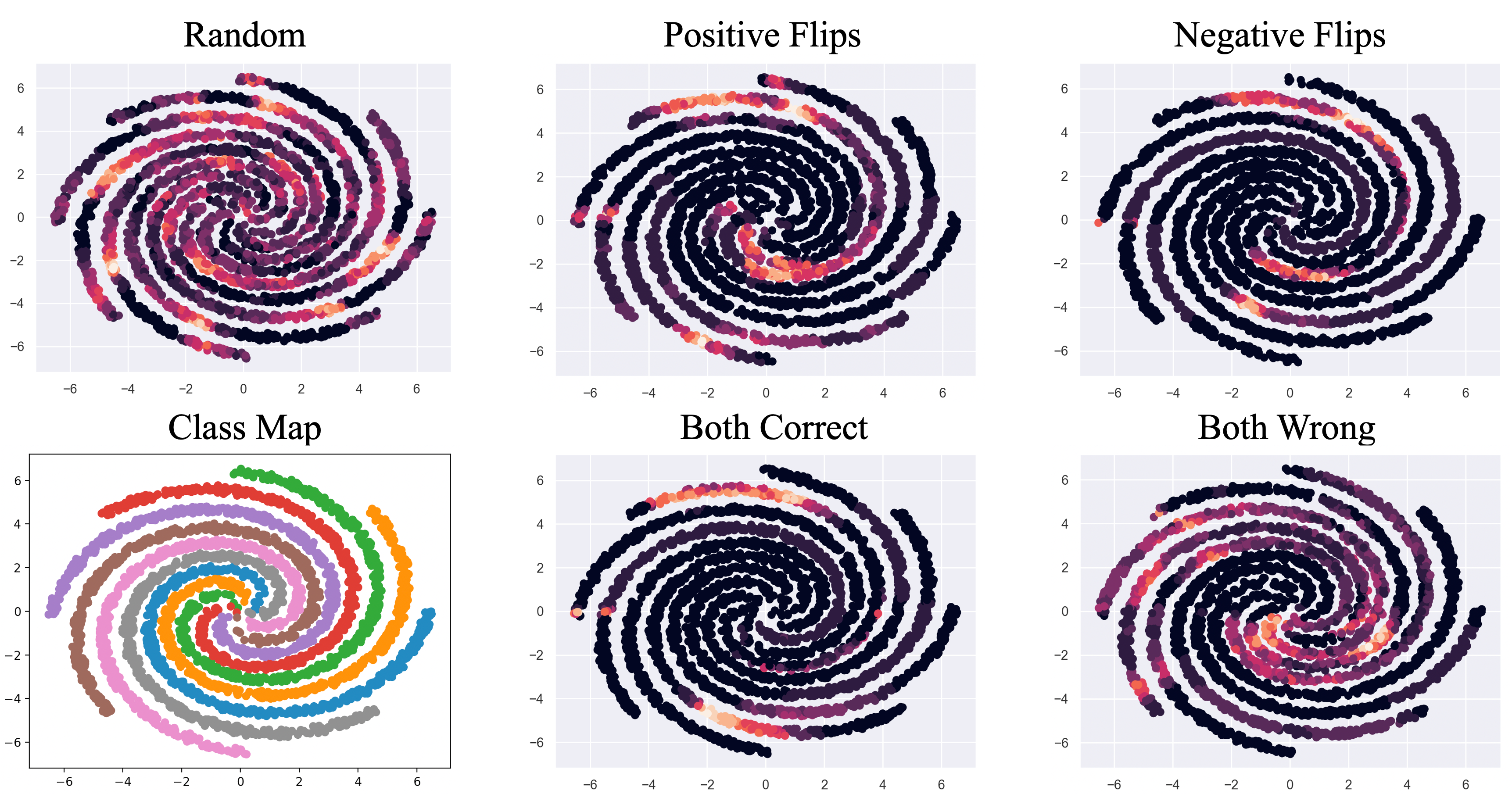}
    \end{center}
    \caption{Negative flips on the 2D artificial spiral dataset. Active learning is performed with random sampling with and without restricting the unlabeled sampling pool. Each plot except the class map represents a different subset restriction. The plot labeled Random has no restriction.}
    \label{fig:artificial-dataset}
\end{figure*}

\section{Discussion}
\label{sec:discussion}
In this section, we investigate what samples frequently constitute negative flips. In Figure~\ref{fig:artificial-dataset}, we visualize negative flips on a 2D artificial dataset with known properties. We consider a spiral dataset with eight different classes. Here, each spiral arm represents a class that starts in the center, and spirals for one full loop of 360°. For each spiral, we generate 500 random samples with noise for our unlabeled training pool resulting in a total of 4000 unlabeled data points. For our test set, we sample the same number of points with the same noise parameters. For our experiments, we train a three-layer MLP architecture over multiple active learning rounds with a random sampling acquisition function and restrict the unlabeled pool to one of five subsets - both correct, both wrong, negative flips, positive flips, and no restrictions (i.e. vanilla random sampling). We start with an initial labeled training set of 100 samples and select 100 samples every active learning round until the entire 4000 data points have been labeled. For each sample within the test set, we log the amount of negative flips over all active learning rounds and show them in the scatter plots in Figure~\ref{fig:artificial-dataset}. Here, lighter red shades indicate a large number of negative flips while darker ones indicate the opposite. We note that regardless which subset we use to restrict the unlabeled pool the proportion of negative flips is reduced to conventional random sampling. In particular, the random plot shows that negative flips occur randomly throughout the test set, regardless of the class or proximity to the center of the spiral. In contrast, sampling from any of the four subsets results in concentrated negative flips. This experiment offers one possible explanation for the increased performance of \texttt{RoSE}. Restricting the unlabeled pool to one of the four subsets can concentrate negative flips into a small subspace with a high density of difficult samples rendering a more consistent model performance throughout the active learning experiment. We further note that the shape and location of regions with high negative flip densities is different depending on the subset used to restrict the unlabeled pool. For instance, sampling from the negative flip subset results in fewer but highly dense negative flip regions while both wrong regions are less dense but encompass more samples. We reason that the both wrong subset contains ambiguous samples that cause the decision boundary to fluctuate more significantly in between active learning rounds. This further supports our choice of utilizing the negative flips subset.


\section{Conclusion}
In this paper, we investigated regression within the context of active learning and found that validation sets can be beneficial for active learning algorithms to improve performance. We provide a detailed analysis where we show the necessity of integrating regression into the active learning objective and further propose \texttt{RoSE} as a plug-in method to improve both accuracy and NFR. A primary observation we made in this paper is that sampling from negative flips is effective in improving performance both in terms of accuracy and software regression. While the empirical performance is impressive investigating the reason still requires significant research effort. Initial investigations in Section~\ref{sec:discussion} suggest that sampling from different subsets compresses the negative flips to smaller regions in the subspace. The observation is compelling but still requires theoretical justification. We see these directions as promising future research.

\bibliographystyle{IEEEtran}
\bibliography{main}

\end{document}